\documentclass[letterpaper, 10pt, conference]{ieeeconf}  

\IEEEoverridecommandlockouts                              

\usepackage[margin=0.75in]{geometry}

\usepackage{graphicx} 

\usepackage{amsmath}
\usepackage{amsfonts}
\usepackage{amssymb}
\usepackage{algorithm}
\usepackage{algpseudocode}
\usepackage{subcaption}
\usepackage[backend=bibtex, sorting=none, style=ieee, doi=false, url=false, eprint=false, isbn=false,maxnames=3,minnames=2]{biblatex}
\usepackage{multirow, booktabs}

\usepackage{makecell}

\usepackage{xcolor}

\title{\LARGE \bf
Spectral Decomposition of Inverse Dynamics for Fast Exploration in Model-Based Manipulation 
}

\author{Solvin Sigurdson$^{1}$, Benjamin Riviere$^{2}$, and Joel Burdick$^{1}$
\thanks{$^{1}$SS and JB are with the Department of Control and Dynamical Systems, California Institute of Technology, Pasadena, CA 91125, USA
        {\tt\small solvin.sigurdson@caltech.edu, jburdick@caltech.edu}}%
\thanks{$^{2}$BR is with the Department of Mechanical and Aerospace Engineering and the Department of Computer Science, New York University, New York City, NY 10012, USA
{\tt\small riviere.b@nyu.edu}}%
}

\begin{document}

\maketitle
\thispagestyle{empty}
\pagestyle{empty}

\begin{abstract}
Planning long duration robotic manipulation sequences is challenging because of the complexity of exploring feasible trajectories through nonlinear contact dynamics and many contact modes. Moreover, this complexity grows with the problem's horizon length. 
We propose a search tree method that generates trajectories using the spectral decomposition of the inverse dynamics equation. This equation maps actuator displacement to object displacement, and its spectrum is efficient for exploration because its components are orthogonal and they approximate the reachable set of the object while remaining dynamically feasible. 
These trajectories can be combined with any search based method, such as Rapidly-Exploring Random Trees (RRT), for long-horizon planning.
Our method performs similarly to recent work in model-based planning for short-horizon tasks, and differentiates itself with its ability to solve long-horizon tasks: whereas existing methods fail, ours can generate 45 second duration, 10+ contact mode plans using 15 seconds of computation, demonstrating real-time capability in highly complex domains.
\end{abstract}

\section{Introduction}

Manipulation is a robot's primary means to interact with and change its environment. However, planning for manipulation is challenging because such interactions have complex dynamics and require creating long-horizon plans that change the state of surrounding objects through a variety of contact modes. In many cases, leveraging an understanding of how the object and the robot can interact with the environment is necessary for successful plans. Consider for instance rearranging objects on a desk where the shape or weight of the objects do not allow for pick and place manipulation, or sliding heavy objects around a construction site.

Model-based optimization has been widely successful in generating real-time trajectory plans for a variety of nonlinear dynamical systems such as drones \cite{scaramuzza2024drones},  spacecraft~\cite{morgan2014model}, and locomotion \cite{ames2016onlinegaits}.
However, applying these methods to robotic manipulation remains a challenge: the change in contact modes during manipulation creates a discontinuity in the gradient of the dynamics, making local linearization and optimization inaccurate. In addition, the number of possible contact modes between manipulators, objects, and the environment grows combinatorially in complex manipulation tasks, inhibiting a direct hierarchical approach of planning discrete contact sequences and then continuous optimization. 

\begin{figure}[thpb]
      \centering
      \includegraphics[scale=0.4]{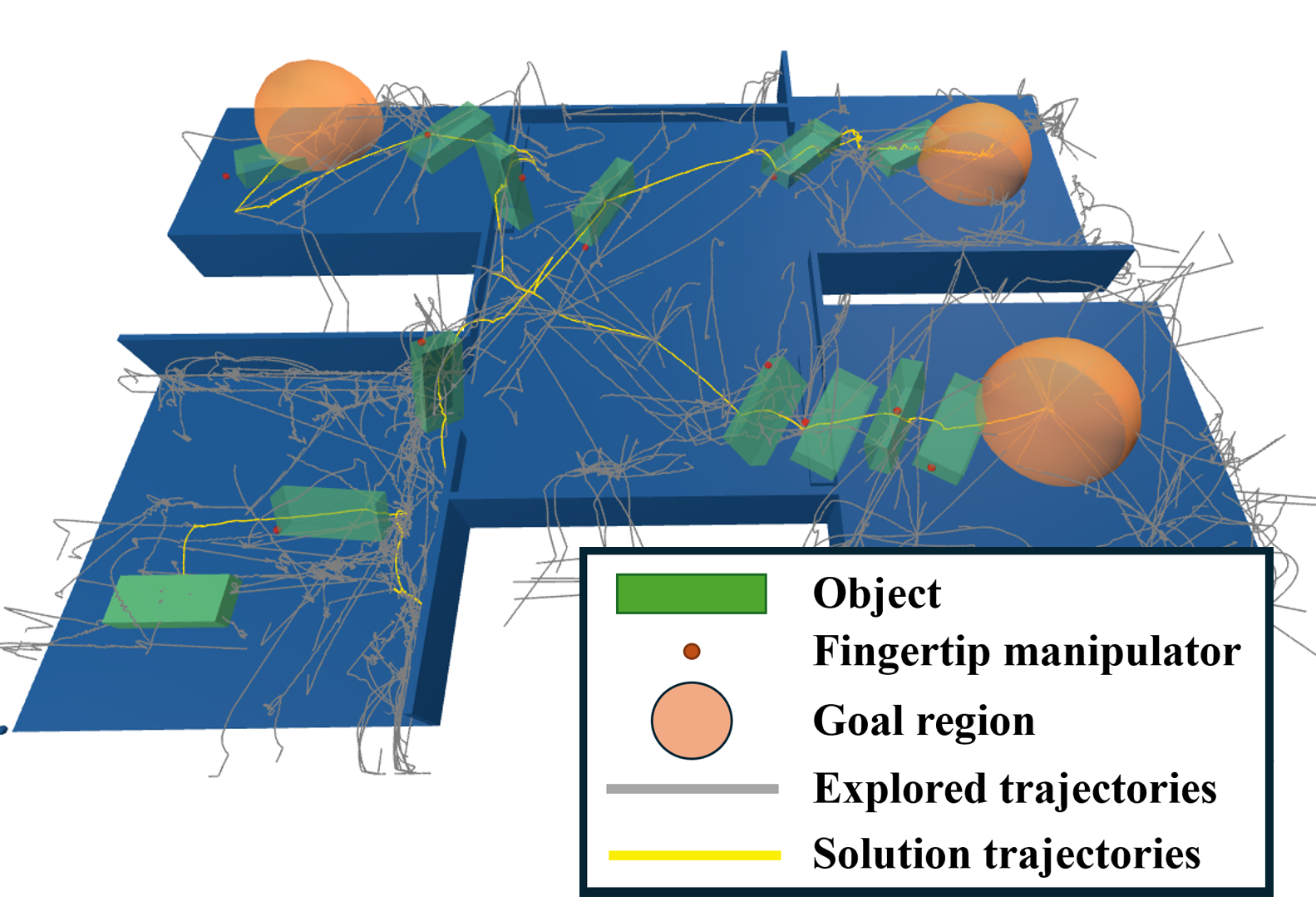}
      \caption{
      Our method efficiently plans manipulation trajectories through several contact modes, discovering behavior like pushing, sliding, rolling, tumbling, and pivoting.  
      In the physics-based simulator, an object (green prism) is moved with non-prehensile manipulation (red dot) through a contact-rich environment to a goal configuration. 
      }
      \label{figs:intropicture}
 \end{figure}

To address these challenges, we propose a new hierarchy: tree search over trajectories generated from the spectrum of the inverse dynamics. 
The algorithm inherits global exploration from search and reduces reliance on inaccurate linearizations and susceptibility to local minima that are both inherent in applying first-order optimization methods to the entire trajectory. 
At the same time, the algorithm uses local dynamics information from the spectrum of the inverse dynamics to reduce search complexity: the local linearization's eigenvectors are efficient for exploration because they are orthogonal and approximate the reachable set of the object while remaining dynamically feasible. 
This spectral representation can be formalized as a Markov Decision Process (MDP), and paired with Rapidly-Exploring Random Trees (RRT)~\cite{lavalle1998rapidly, karaman2011sampling} (or other search algorithms) to chain branches together and efficiently construct long-horizon trajectories.

We verify this approach in simulation, see Fig.~\ref{figs:intropicture}, on non-prehensile manipulation tasks that slide, pivot, and rotate a rectangular prism through a highly varied environment involving many contact modes. Interacting with the environment via non-trivial extrinsic dexterity is necessary in order to complete the task. 
On a desktop computer, our method uses $\sim$15 seconds of computation time to generate plans of 45 seconds physical duration, demonstrating long-horizon planning capability while remaining fast enough for real-time deployment and re-planning. 

Overall, our method produces real-time complex plans involving contact between objects, manipulators, and the surrounding environment. The method is model-based and can be applied zero-shot without pre-training, fine-tuning or hyperparameter tuning in a variety of scenarios.
Although the focus of this work is extrinsic dexterity and non-prehensile manipulation, it could be extended in the future to dexterous manipulation and loco-manipulation. 

\section{Background and Related Works}

Robot manipulation is a rich field that spans hybrid systems theory, optimization, search, and machine learning. 

The problem of planning and controlling an object's motion via pushing, sliding, rolling, tumbling, pivoting, etc. was traditionally viewed as a hybrid system of discrete contact modes, where within each contact mode the model reduces to a continuous non-linear system to which standard optimization techniques can be applied \cite{feedbackcontrolhybridsystems}. 
Although this formulation can be solved with mixed-integer-programming methods~\cite{marcucci2019mipaffinesystems}~\cite{marcucci2020warmstartmip} or search over contact mode transitions~\cite{toussaint2018tamp}~\cite{kingston2023scalingmultimodalplanning}, the complexity of these algorithms scales with the number of contact modes, limiting this formulation for long-horizon manipulation tasks.

To avoid contact mode enumeration, contact-implicit methods use compliant contact models to generate contact forces based solely on the relative configuration of objects, leading to a single unified dynamics model for all contact modes ~\cite{posa2014cito}~\cite{lecleach2024cimpc}. 
This approach enables model-based optimization methods to generate solution trajectories. However the approach is heavily susceptible to local minima, and it requires a good initial guess~\cite{kurtz2023inverse}.

Sampling-based methods are a standard approach to overcome local minima. For example, one could sample the input space directly at each timestep using Predictive Sampling ~\cite{howell2022predictivesamplingrealtimebehaviour}, but sampling in a high-dimensional space quickly becomes intractable for long-horizon plans whose intermediate goals do not lie directly between the start and goal.

It is possible to reduce the search complexity by focusing on the object's motion alone~\cite{cheng2024mcts}~\cite{pang2023rrt}~\cite{zhu2023efficient}. However, efficiently proposing dynamically realizable motions for the object is challenging: ~\cite{cheng2024mcts} uses a nested search structure to find dynamically feasible motions, which can lead to inefficiency if inner searches fail to produce solutions for outer layers, ~\cite{pang2023rrt} does not address uni-directional property of friction and is limited to a convex relaxation of the contact dynamics, and ~\cite{suh2025dexterouscontactrichmanipulationcontact} proposes dynamically feasible paths but requires time to first build a roadmap offline.

Our method falls into this category of search-based methods and achieves both low sample complexity and dynamic feasibility by using the spectral decomposition of the inverse dynamics equation: its local linearization's eigenvectors are efficient for exploration because they are orthogonal and approximate the reachable set of the object while remaining dynamically feasible. 
Both our approach and~\cite{riviere2024sets} use spectral decompositions to reduce search complexity. However, whereas they use the spectrum of the locally linearized system's Grammian, we use the spectrum of the inverse dynamics equation, which is a specialized transformation that focuses on actuator to object displacement. 

Recent progress has been made using data-driven methods. 
End-to-end visuomotor policies grounded in reinforcement learning and behaviour cloning can directly map camera inputs to manipulator actions, removing the need to explicitly plan through challenging contact dynamics \cite{levine2016endtoendtrainingdeepvisuomotor}. 
The main challenge for these methods is demonstrating appropriate generalization and reliability, as success on the task often depends on its proximity to training examples, although recent work in foundation models for robotics are attempting to overcome this difficulty \cite{pizero2024}. An alternate line of work considers how to use generative models to produce trajectories directly \cite{kurtz2025generativepredictivecontrolflow} although this again suffers from canonical out-of-distribution (OOD) challenges if the test time data distribution is not similar enough to that of the training. 
In contrast, purely model-based methods are inherently general because they do not rely on a training distribution, and are more interpretable, which makes them more reliable in an industrial setting. 
However, they do still have challenges: model-based methods suffer when the dynamics model is inaccurate, requiring good feedback controllers or online adaptation to bridge the sim-to-real gap. 
 
Some works attempt to balance these regimes by merging physics-driven models with data- and simulation-driven adaptation. 
For instance, model-based methods can generate inductive biases for learning based methods, achieving the best of both worlds~\cite{zhu2025learningfrommodels, riviere2021neural}.

\section{Method}

\subsection{Problem Definition}

\begin{figure}[b]
      \centering
      \includegraphics[scale=0.45]{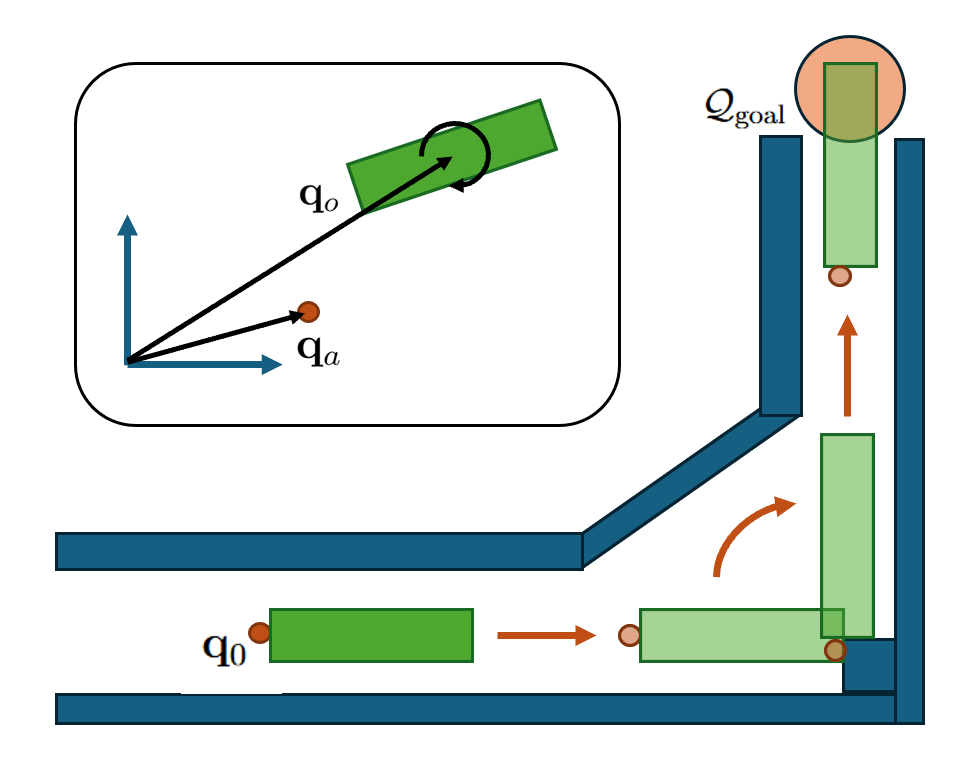}
      \caption{Top left: Variables for the problem setup. Bottom right: Simple example of the fingertip exhibiting extrinsic dexterity in order to move an object through a maze.
      }
      \label{fig:problem_defn}
\end{figure}

This work seeks to efficiently address the problem of planning the motions of a single rigid body over a complex geometric terrain, where a high number of contact interactions between the rigid body, the manipulator, and the surrounding environment may be needed to reach the goal configuration. The geometry and basic material properties of the body, manipulator, and environment are assumed to be known a priori. For the purpose of simplicity in exposition, we first consider the manipulator as a single spherical fingertip, although the algorithm extends to an arbitrary number of fingertips, which could be remapped to a dexterous hand or bimanual manipulator. 

Taking $\mathbf{q} = \begin{bmatrix} \mathbf{q}_{a} \; \mathbf{q}_o\end{bmatrix} \in \mathbb{R}^{n_a + n_o} = \mathbb{R}^{n}$ as the combined configuration of the $n_a$ actuated degrees of freedom ($\mathbf{q}_{a}$) and $n_o$ object degrees of freedom ($\mathbf{q}_o$), the problem is to find a dynamically feasible trajectory $\mathbf{q}_{0:H} \in \mathbb{R}^{n \times H}$ where $\mathbf{q}_0 = \mathbf{q}_\text{init}$ and $\mathbf{q}_H \in \mathcal{Q}_\text{goal}$ for a given $\mathbf{q}_\text{init}$ and $\mathcal{Q}_\text{goal}$, and $H$ is the length of the trajectory (see Fig. \ref{fig:problem_defn}). 
A trajectory is dynamically feasible if it is generated by a physics-based simulator -- we use Drake~\cite{drake}, which implements hydroelastic contact models and is considered a close approximation of real world contact dynamics.

This problem setting is particularly applicable to cases where the manipulator does not have complete actuation authority over the object's degrees of freedom (e.g. non-prehensile manipulation). In this setting, the process of changing the object's state often relies on interactions between the object and the environment. Note that the framework is not limited to this case, and in fully actuated circumstances (i.e. where classical pick and place grasping is available), the method can discover grasping solutions. In this work, $n_a = 3$ and $n_o = 6$ for a single fingertip and a single free rigid body with Euler angle parameterization.

\subsection{Preliminaries}

We model the dynamics of the object and manipulator using the standard Euler-Lagrange equations, with a term for the contact forces $\mathbf{\lambda}_i$ between the $i^\text{th}$ object pair. 

\begin{align}
    \mathbf{M}(\mathbf{q})\ddot{\mathbf{q}} + \mathbf{k}(\mathbf{q}, \dot{\mathbf{q}}) = \mathbf{B}\mathbf{u} + \sum_i\mathbf{J}_i(\mathbf{q})^{\top} \mathbf{\lambda}_i
    \label{eq:dynamics}
\end{align}
where $\mathbf{M} \in \mathbb{R}^{n \times n}$ is the mass matrix, $\mathbf{k}(\mathbf{q}, \dot{\mathbf{q}})$ accounts for gravity and Coriolis terms, $\mathbf{B} \in \mathbb{R}^{n \times n_a}$ is the actuation matrix, $\mathbf{u} \in \mathbb{R}^{n_a}$ are the forces applied to the actuated joints, and $\mathbf{J}_i(\mathbf{q}) \in \mathbb{R}^{3 \times n}$ are the Jacobians mapping the contact forces into the generalized variables. We use the compliant contact model from \cite{kurtz2023inverse}\cite{huntcrossley}, which uniquely specifies the contact force $\lambda_i \in \mathbb{R}^3$ between objects as a function of configuration $\mathbf{q}$. 
Drake is used to calculate queries for relative separation and normal velocity in the contact model.

Similar to ~\cite{kurtz2023inverse}, given $\mathbf{q},\dot{\mathbf{q}},\ddot{\mathbf{q}}$ we can invert the dynamics to find the instantaneous external forces $\mathbf{\tau}_\text{ext} \in \mathbb{R}^n$ that would result in $\mathbf{q},\dot{\mathbf{q}},\ddot{\mathbf{q}}$ assuming full actuation authority:   
\begin{align}
    \mathbf{\tau}_\text{ext}(\mathbf{q}, \dot{\mathbf{q}}, \ddot{\mathbf{q}}) = \mathbf{M}(\mathbf{q})\ddot{\mathbf{q}} + \mathbf{k}(\mathbf{q}, \dot{\mathbf{q}}) - \sum_i\mathbf{J}_i(\mathbf{q})^{\top} \mathbf{\lambda}_i(\mathbf{q})
\end{align}
However, our system is underactuated and cannot apply forces to all of the degrees of freedom, therefore a given $\mathbf{q},\dot{\mathbf{q}},\ddot{\mathbf{q}}$ is only feasible if the necessary external forces on the unactuated degrees of freedom are zero: 
\begin{equation}
    \mathbf{h}(\mathbf{q}, \dot{\mathbf{q}}, \ddot{\mathbf{q}}) = \mathbf{H}\mathbf{\tau}_\text{ext}(\mathbf{q}, \dot{\mathbf{q}}, \ddot{\mathbf{q}}) = \mathbf{0}
    \label{eq:implicit_dynamics_eqn}
\end{equation}
where $\mathbf{H} \in \mathbb{R}^{n_o\times n} $ selects the unactuated elements of $\mathbf{\tau}_\text{ext}$. 

\subsection{Spectral Decomposition of Inverse Dynamics Equation}

Given an initial configuration, $\mathbf{q}_0$, we want to compute a set of trajectories that are representative of the possible motions of the object. 
These trajectories will be used later in the search methods. 

First, we approximate $\mathbf{h}$ with a linearization $\hat{\mathbf{h}}$, about a nominal point: $(\bar{\mathbf{q}}, \dot{\bar{\mathbf{q}}}, \ddot{\bar{\mathbf{q}}})$:
\begin{align}
    \label{eq:linearized_dynamic_feasibility}
    \hat{\mathbf{h}}(\mathbf{q}, \dot{\mathbf{q}}, \ddot{\mathbf{q}}) 
    &= \mathbf{h}(\bar{\mathbf{q}}, \dot{\bar{\mathbf{q}}}, \ddot{\bar{\mathbf{q}}}) 
    + \frac{\partial\mathbf{h}}{\partial\mathbf{q}}(\mathbf{q} - \bar{\mathbf{q}}) \nonumber \\ 
    & \quad + \frac{\partial\mathbf{h}}{\partial\mathbf{\dot{q}}} (\mathbf{\dot{q}} - \dot{\bar{\mathbf{q}}})
    + \frac{\partial\mathbf{h}}{\partial\mathbf{\ddot{q}}} (\mathbf{\ddot{q}} - \ddot{\bar{\mathbf{q}}})
\end{align}

We simplify this function by only considering a subset of possible trajectories. 
In particular, we fix the linearization trajectory at the current position with zero velocity and zero acceleration: $(\bar{\mathbf{q}}, \dot{\bar{\mathbf{q}}}, \ddot{\bar{\mathbf{q}}}) = (\mathbf{q}_0, \mathbf{0}, \mathbf{0})$, and we assume that a constant acceleration occurs over timestep $\Delta t$ to achieve velocity $\dot{\mathbf{q}}$: $(\mathbf{q}, \dot{\mathbf{q}}, \ddot{\mathbf{q}}) = (\mathbf{q}_0, \dot{\mathbf{q}}, \dot{\mathbf{q}}/\Delta t)$. 
Although these assumptions limit the set of trajectories we can consider, they are reasonable for small $\Delta t$ in cases when the state of the system comes to rest when changing manipulator positions. 

Substituting in these assumptions, $\hat{\mathbf{h}}(\cdot, \cdot, \cdot)$ is simplified to: 
\begin{align}
    \hat{\mathbf{h}}(\mathbf{q}, \dot{\mathbf{q}}, \ddot{\mathbf{q}}) 
    &= \left( 
        \frac{\partial\mathbf{h}}{\partial\mathbf{\dot{q}}} + 
        \frac{\partial\mathbf{h}}{\partial\mathbf{\ddot{q}}} 
        \frac{1}{\Delta t} 
    \right) 
    \mathbf{\dot{q}}
\end{align}
where we assume $\mathbf{h}(\mathbf{q}_0, \mathbf{0}, \mathbf{0}) = \mathbf{0}$ because the system is linearized about a dynamically feasible point. 

Setting our approximation of $\mathbf{h}$ to zero and applying chain rule for $\frac{\partial \mathbf{h}}{\partial \mathbf{\ddot{q}}}$ leads to the following equation: 

\begin{equation}
    \frac{\partial\mathbf{h}}{\partial\dot{\mathbf{q}}}\dot{\mathbf{q}} = \begin{bmatrix}
        \frac{\partial\mathbf{h}}{\partial\dot{\mathbf{q}}_a} \; \frac{\partial\mathbf{h}}{\partial\dot{\mathbf{q}}_o}
    \end{bmatrix}
    \begin{bmatrix}
        \dot{\mathbf{q}}_a \\
        \dot{\mathbf{q}}_o
    \end{bmatrix} = \mathbf{0}
\end{equation}
which can be manipulated into the desired form:
\begin{equation}
    \dot{\mathbf{q}}_o = -\left(\frac{\partial\mathbf{h}}{\partial\dot{\mathbf{q}}_o}\right)^{\dagger}\frac{\partial\mathbf{h}}{\partial\dot{\mathbf{q}}_a}\dot{\mathbf{q}}_a = \mathbf{A}(\mathbf{q}_a, \mathbf{q}_o) \dot{\mathbf{q}}_a
    \label{eq:pseudodynamical system}
\end{equation}
where $\dagger$ is the pseudoinverse. 

The configuration-dependent matrix $\mathbf{A}(\mathbf{q}) \in \mathbb{R}^{n_o \times n_a}$ is a mapping between the motion of the actuated degrees of freedom and the object motion. 
The range space of $\mathbf{A}$ is the set of possible velocities for the object given the current state $\mathbf{q}_a, \mathbf{q}_o$, which cover the one step reachable set when integrated. 
We can approximate this set with a finite number of vectors constructed from the eigenvectors $\mathbf{v}_i$ of $\mathbf{A}\mathbf{A}^\top$:
\begin{align}
    \sigma_i^2, \mathbf{v}_i = \text{Eig}(\mathbf{A}\mathbf{A}^\top)
    \label{equations:eigen}
\end{align}

In implementation, we calculate derivatives $\frac{\partial\mathbf{h}}{\partial\dot{\mathbf{q}}_a}$ and $\frac{\partial\mathbf{h}}{\partial\dot{\mathbf{q}}_o}$ around nominal point $(\bar{\mathbf{q}}, \dot{\bar{\mathbf{q}}}, \ddot{\bar{\mathbf{q}}})$ by finite differences, and ignore eigenpairs with eigenvalues such that $\frac{\sigma_i^2}{\sigma_\text{max}^2} < C_\text{eig}$.

\subsection{Reachable Set Approximation}

We use the spectral decomposition from~\eqref{equations:eigen} to construct a set of candidate trajectories with Algorithm~\ref{alg:ReachableSet}, which is outlined in the following steps.

First, we sample $N_\text{grasps}$-many initial configurations for the same object state $\mathbf{q}_o$ by sampling fingertip locations on the object ($\text{ObjGeometry}$) and using inverse kinematics to determine associated configurations $\mathbf{q}_j$ (Lines 1 and 2 of Algorithm~\ref{alg:ReachableSet}). The linearized object motion directions $\mathbf{v}_i$ are computed using the spectral decomposition from~\eqref{equations:eigen} for each initial configuration (Line 3 of Algorithm~\ref{alg:ReachableSet}). 

Some of these motions will be infeasible for the nonlinear system because of the linearization error. For example, linearized dynamics allow the fingertips to pull on the object instead of push, and for environment penetration by the object. We filter out the former infeasible proposals by comparing the fingertip velocity vector with the fingertip-to-object normal vectors (removed if inner product below threshold $C_\text{fingertip}$), and the latter by comparing any significant object center of mass velocity with environment-to-object normals (removed if inner product below $C_\text{env}$ and translational to rotational motion ratio over $C_\text{ratio}$). 
This procedure is denoted as \textsc{FilterProposals} (Line 4 of Algorithm~\ref{alg:ReachableSet}).

At this point, the number of representative motions is proportional to $N_\text{grasps}$. However, these motions could be similar to each other, and each additional motion costs computation by increasing the branching factor of the resulting tree.  
To maintain an efficient representation, a representative set of motions is chosen using kmeans clustering with $N_\text{clusters}$ (Line 5 of Algorithm~\ref{alg:ReachableSet}). 
The cluster centers in $\mathcal{V_\text{cc}}$ are then normalized by matrix $\mathbf{W} \in \mathbb{R}^{n_o \times n_o}$ to balance translational and rotational degrees of freedom (Line 6 of Algorithm~\ref{alg:ReachableSet}).

\begin{figure}[thpb]
      \centering
      \includegraphics[scale=0.7, trim={0cm 1.5cm 0cm 1.5cm}, clip]{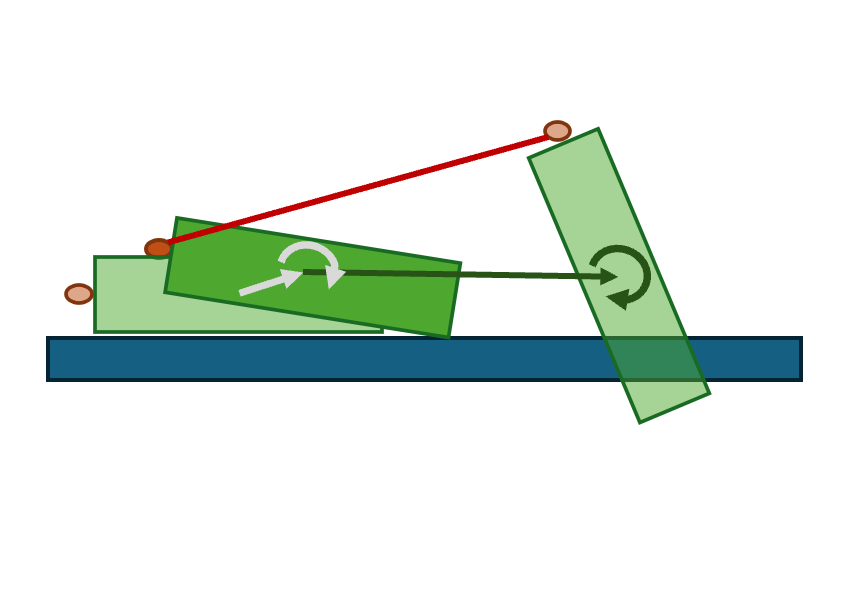}
      \caption{
      An illustration of the fingertip controller. The dark green arrows are the linearized motion $\mathbf{v}_i$ projected forward $T_\text{proj}$ seconds, the gray arrows are the nonlinear motion, and the end of the red line is the fingertip reference point.} 
      \label{figs:fingertip_controller}
 \end{figure}

Given the linearized object velocity from the prior steps, we now compute a trajectory $\mathbf{q}_\text{traj}^i \in \mathbb{R}^{n \times N_i}$ using the exact nonlinear dynamics, encapsulated in $\textsc{PDRollout}$ (Line 7 of Algorithm~\ref{alg:ReachableSet}). 
To achieve this, we use a proportional derivative (PD) controller that actuates the fingertips and tracks a desired position trajectory that keeps the fingertip stationary relative to the object's surface, while the object moves with velocity $\mathbf{v}_i$, see Fig. \ref{figs:fingertip_controller}. 
The desired fingertip position setpoint for the PD controller can be calculated at any time $t_0$ using Equations \ref{eq:projection1}-\ref{eq:pd_controller} evaluated at time $t_0 + T_\text{proj}$. 

\begin{align}
    \mathbf{q}_{o}(t) &= \mathbf{q}_{o}(t_0) \ + \mathbf{v}_i (t - t_0) 
    \label{eq:projection1} \\
    \bar{\mathbf{p}}_w(t) &= \mathbf{T}_{wo}({q}_{o}, t)\bar{\mathbf{p}}_o \label{eq:projection2} \\
    \mathbf{u} &= \mathbf{K}_d\dot{\mathbf{q}}_a + \mathbf{K}_p(\mathbf{q}_a - \mathbf{p}_w)
    \label{eq:pd_controller}
\end{align}
where $T_\text{proj}$ is the duration of the forward projection in seconds, $\bar{\mathbf{p}}_o \in \mathbb{R}^4$ is the homogeneous vector associated with $\mathbf{p}_o \in \mathbb{R}^3$ the contact location in the object body frame, $\mathbf{T}_{wo} \in \mathbb{R}^{4 \times 4}$ is the  transform to the world frame, and $\bar{\mathbf{p}}_w \in \mathbb{R}^4$ is the homogeneous vector associated with $\mathbf{p}_w \in \mathbb{R}^3$ the contact location in the world frame. 
The setpoint is updated every $T_\text{track}$ seconds during the simulation rollout using the current object position and fingertip contact location, which may have shifted if the finger slips during the rollout. This controller explores object motion near $\mathbf{v}_i$  while accounting for non-linear contact dynamics via the simulator. 

\begin{algorithm}
\caption{}\label{alg:ReachableSet}
\begin{algorithmic}
\State \textbf{Input}: Configuration $\mathbf{q}_o$
\State \textbf{Output}: Reachable set $\mathcal{R}$
\State $\textbf{Hyperparameters} \ N_\text{grasps}$, $\mathbf{W}$ 
\end{algorithmic}
\begin{algorithmic}[1]
\State $\mathcal{P} \gets \{\mathbf{p}_j^{o} \in \mathbb{R}^3 |\mathbf{p}_j^{o} \sim \text{ObjGeometry}, j=1 \dots N_\text{grasps}\}$
\State $\mathcal{Q} \gets \{\mathbf{q}_j = \text{InverseKinematics}(\mathbf{p}_j^o), \forall j=1 \dots N_\text{grasps}\}$
\State $\mathcal{V}_\text{full} \gets \{\mathbf{v}_i |\ \_, \mathbf{v}_i = \text{Eig}(\mathbf{A}_j\mathbf{A}_j^\top), \ j = 1 ...N_\text{grasps} \}$
\State $\mathcal{V}_\text{feasible} = \textsc{FilterProposals}(\mathcal{V}_\text{full})$
\State $\mathcal{V}_\text{cc} \gets \textsc{Kmeans}(\mathcal{V}_\text{feasible})$
\State $\mathcal{V}_\text{norm} \gets \{\mathbf{v}_i' \ | \ \mathbf{v}_i'=\frac{\mathbf{v}_i}{ \sqrt{\mathbf{v}_i^T \mathbf{W} \mathbf{v}_i} } \ \forall \mathbf{v}_i \in \mathcal{V}_{cc} \}$
\State $\mathcal{R} \gets \{\mathbf{q}_\text{traj}^i | \mathbf{q}_\text{traj}^i = \textsc{PDRollout}(\mathbf{q}_i, \mathbf{v}_i) \ \forall \mathbf{v}_i \in \mathcal{V}_\text{norm}\}$
\newline
\Return $\mathcal{R}$
\end{algorithmic}
\end{algorithm}

\begin{figure}[thpb]
      \centering

      \includegraphics[scale=0.45]{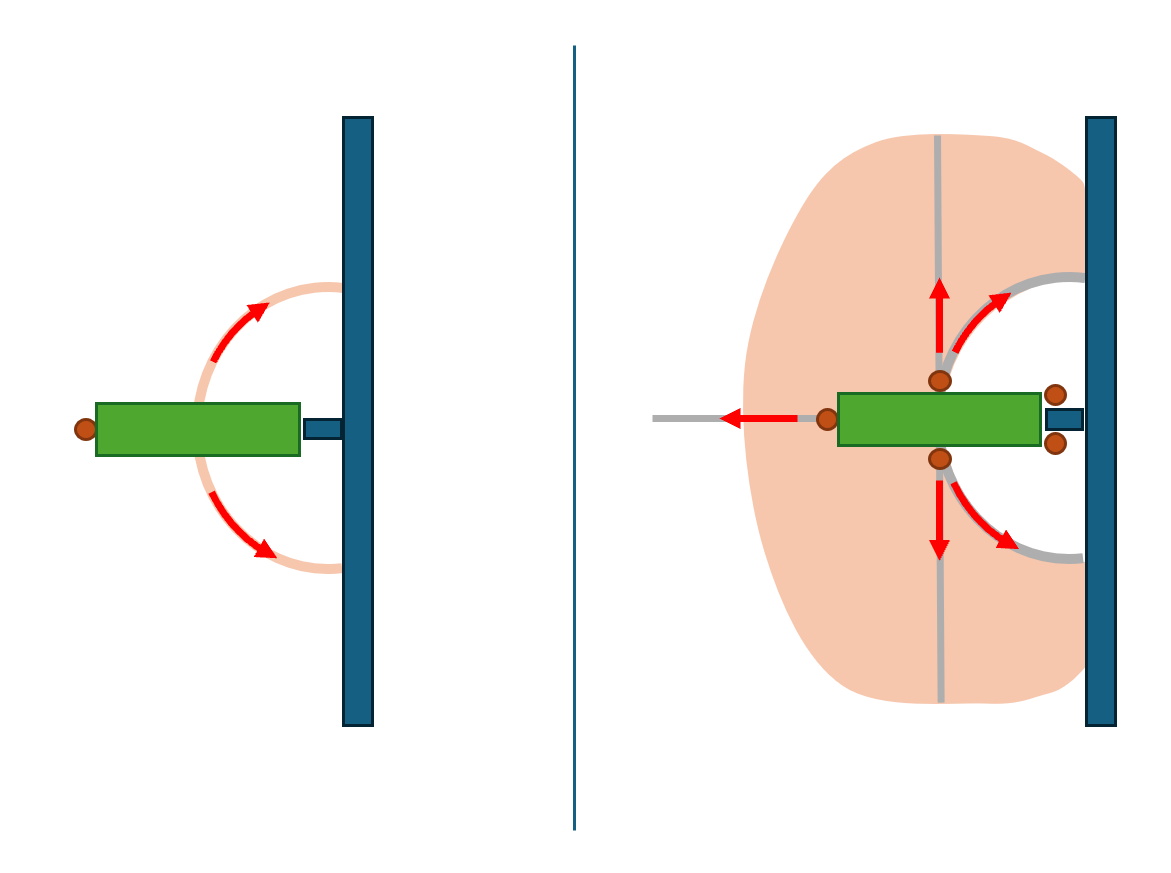}
      \caption{Left: For a single configuration where the object is pinned between the fingertip and the environment, proposed motion directions $\mathbf{v}_i$ are shows with red arrows, while the true reachable set for the center of the box is shown in pink. 
      Right: Same as the left image except for a set of 5 different initial configurations to capture sampling different contacts between the object and manipulator. The approximate reachable set for the object's COM is shown in gray, and efficiently captures a distinct subset of all possible motions in pink.}
      \label{figs:reachable_sets}
 \end{figure}

While the trajectory is being rolled out, we check if the object loses contact with the fingertip over threshold $d_\text{contact}$, whether motion has stopped below threshold $v_\text{stopped}$ relative to the initial velocity, whether the object has rotated by $\phi_\text{max}$, or whether time $T_\text{max}$ has elapsed, in which case we stop tracking the proposed motion $\mathbf{v}_i$. 
The approximate reachable set $\mathcal{R}$ is the union of the trajectories produced by the rollouts, see Fig. \ref{figs:reachable_sets}. 
We specify the hyperparameters of the subroutines \textsc{PDRollout}, \textsc{FilterProposals}, and \textsc{Kmeans} in Sec.~\ref{sec:experiment}.

\subsection{Reachable Sets in Search-Based Planners}
\begin{figure}[thpb]
      \centering
      \includegraphics[scale=0.4]{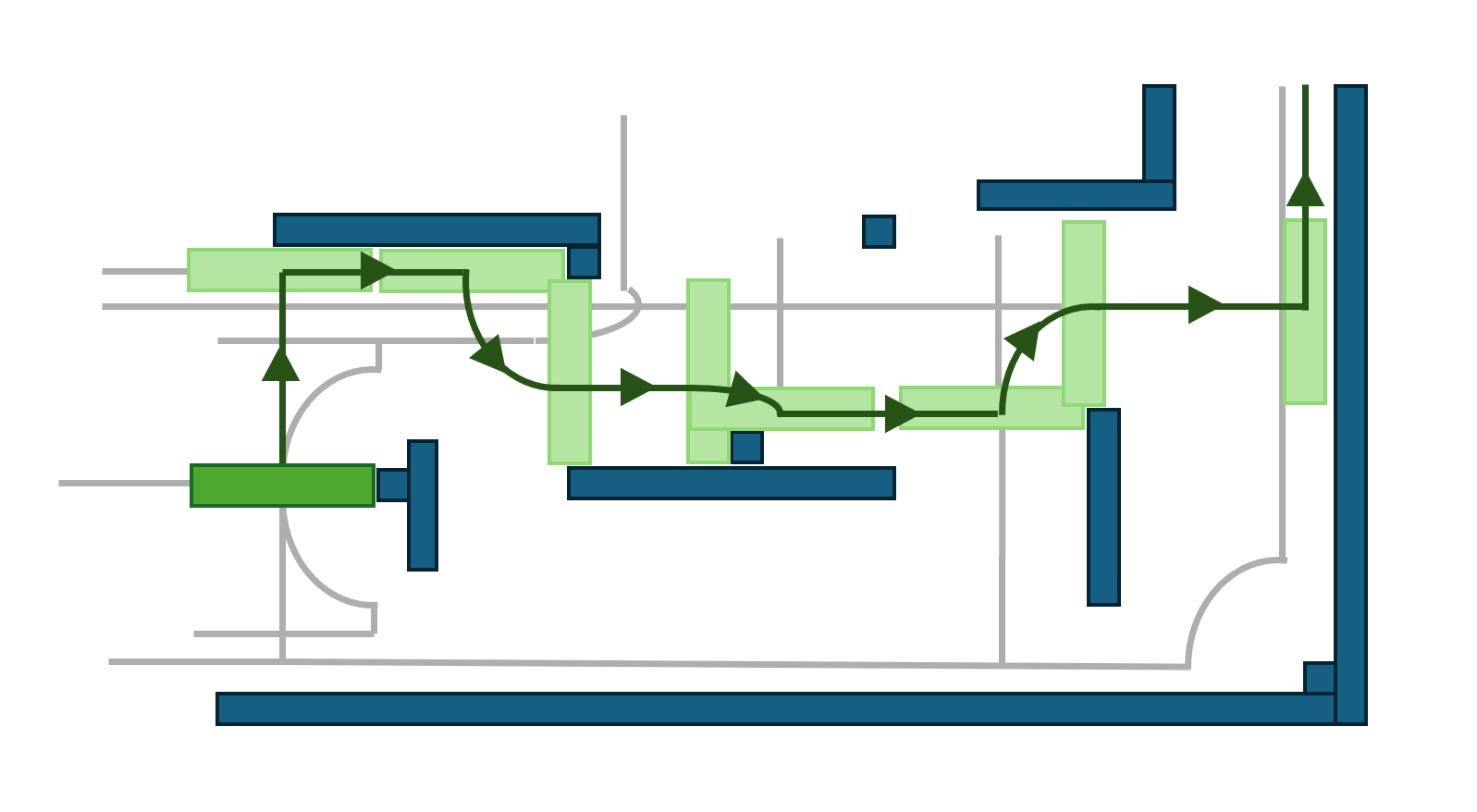}
      \caption{Pictoral representation of the MDP tree for a simple maze problem. In grey are the possible motion paths for the center of mass, while in dark green is a possible path chosen by the RRT algorithm.}
      \label{figs:RRT_expansion}
 \end{figure}

We use the reachable set to construct a Markov Decision Process (MDP) that can be used in search-based planners, see Fig. \ref{figs:RRT_expansion}, and we specify one possible algorithmic variant using Rapidly-Exploring Random Trees (RRT). 

We define the elements of the MDP, the novelty of which is the action set generator that uses our previously defined reachable set $\mathcal{R}$: 
\begin{align}
    \label{eq:mdp_definition}
    &\mathcal{S}: \{\mathbf{q}_{0:H} \in \mathbb{R}^{n \times H} ,\; H \in \mathbb{N}\} \\
    \label{eq:mdp_action}
    &\mathcal{A}(\mathbf{s} \in \mathcal{S}): \{\mathbf{q}_{0:N_i}^i | \mathbf{q}_{0:N_i}^i \in \mathcal{R}(\mathbf{q}_{o, H})\} \\
    \label{eq:mdp_transition}
    &\mathcal{T}(\mathbf{s} \in \mathcal{S}, \mathbf{a} \in \mathcal{A}): \mathbf{s}’ = \mathbf{s} \oplus \mathbf{a}\\
    \label{eq:mdp_reward}
    &\mathcal{R}(\mathbf{s} \in \mathcal{S}, \mathbf{a} \in \mathcal{A}, \mathbf{s}’ \in \mathcal{S}): \{1 \text{ if } \mathbf{s}’_{H'} \in \mathcal{Q}_\text{goal}\}
\end{align}

where the $\mathbf{q}_{0:H}$ are discretized system trajectories over a horizon $H$, which grows as new path segments are added, $N_i$ is the length of the $i^\text{th}$ trajectory in the reachable set, 
$\oplus$ denotes concatenation of the trajectories in time,
$\mathcal{Q}_\text{goal}$ is the set of states that are considered sufficiently close to the goal state to end the search (with radius $R_\text{terminal}$), 
and the reward is not used in every search method (e.g. RRT), but is included here for completeness.

We implement a standard RRT algorithm~\cite{lavalle1998rapidly, karaman2011sampling} with the following choices of \textbf{Sample}, \textbf{Select}, and \textbf{Steer}. 
The choices of \textbf{Sample} and \textbf{Select} are standard, our novelty is in the construction of $\mathcal{A}(s^*)$ that is used to \textbf{Steer}. 

\textbf{Sample}: Randomly sample an object configuration within a predefined problem-dependent bounding box: $\mathbf{q}_\text{sub-goal}^{o}~\sim~[a_i, b_i]^{n_o}$. 
We use standard "goal biasing": we sample the center of the problem goal region $\mathcal{Q}_\text{goal}$ with frequency $\alpha \in [0,1]$. 

\textbf{Select}: We select node $\mathbf{s}^*$ for expansion, which has the smallest Euclidean distance between the trajectory's end state $\mathbf{q}_H$ and the randomly sampled configuration. 
While the distance calculation can include both translation and orientation factors, we used only a translation-based metric.

\textbf{Steer}: Compute the transition for a random action $\mathbf{a} \in \mathcal{A}(\mathbf{s}^*)$ from Algorithm \ref{alg:ReachableSet}, and add $N_\text{nodes}$ new states to the tree, evenly spaced along the new trajectory segment $\mathbf{a}$. 

\vspace{0.3cm}
\textit{On Algorithmic Completeness} - Due to environment contact constraints, solution trajectories may exist in narrow passages of configuration space. We suspect our algorithm is resolution complete through these passages as individual rollout length decreases, although this increases computational cost and we do not include a formal proof.

\section{Experimental Results and Discussion}
\label{sec:experiment}

We evaluate our method's efficiency and accuracy on several simulations, focusing on non-prehensile manipulation and extrinsic dexterity. 
The computation was entirely CPU-based (Intel i9-14900K 3.20 GHz CPU) since forward rollouts are done in Drake. The forward rollouts were parallelized on all 24 cores.
In order to avoid instantaneous changes of fingertip position (recalling each segment in $\mathcal{A}$ in (\ref{eq:mdp_action}) begins with the single fingertip in a new position), we additionally append $\mathbf{q}_\text{settle}$ to each transition in (\ref{eq:mdp_transition}), which is a short forward simulation with the fingertip removed. Forward simulation is done with discrete timestep $\Delta t_\text{action}$ during computation of elements of $\mathcal{R}$, and $\Delta t_\text{settle}$ during fingertip removal.

We use the same hyperparameters for all experiments: $N_\text{grasps} = 100$, $N_\text{max}= 50$, $T_\text{proj} = 2.0$ [s], $T_\text{track} = 0.4$ [s], $\alpha = 0.2$, $K_p = 5000$, $K_d = 500$, $\Delta t_\text{action} = 10^{-2}$ [s], $\Delta t_\text{settle} = 10^{-3}$ [s], $\Delta t = 10^{-3}$ [s], $d_\text{contact} = 10^{-3}$ [m], $v_\text{stopped} = 10^{-2}$, $\phi_\text{max} = 2.5$ [rad], $T_\text{max} = 10$ [s], $N_\text{clusters} = 9$, $C_\text{fingertip} = -0.1$, $C_\text{env} = -0.3$, $C_\text{ratio} = 0.05$, $C_\text{eig} = 10^{-6}$, $R_\text{terminal} = 0.2$ [m], $N_\text{nodes} = 5$, $\mathbf{W}$ has diagonal elements 40.5 for translational d.o.fs and 0.405 for rotational d.o.fs - accounting for an object length of roughly 0.1m and normalizing a unit of rotation to $\pi/2$. Note that if the object properties varied more significantly, the impedance gains and length scales could be automatically chosen from the object's inertia and size.

\subsection{Short-Horizon Extrinsic Dexterity Tasks}

\begin{figure}
    \centering
  \begin{subfigure}[t]{0.31\linewidth}\centering
    \includegraphics[trim={2 0 2 0}, clip, width=.99\linewidth]{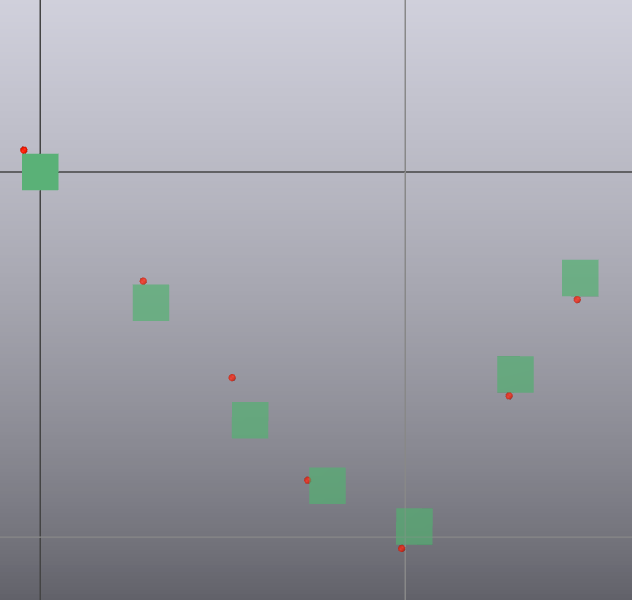}
    \caption{Planar Pusher}\label{fig:simple_task_planar_pusher}
  \end{subfigure}
  \begin{subfigure}[t]{0.31\linewidth}\centering
    \includegraphics[trim={0 0 0 2}, clip, width=.99\linewidth]{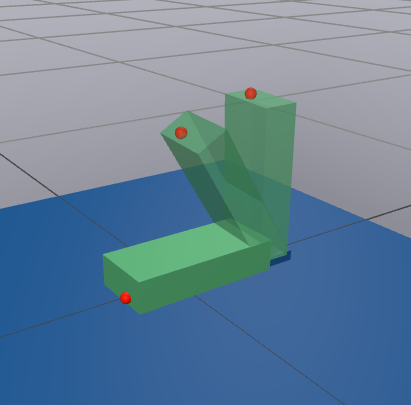}
    \caption{Card Flip}\label{fig:simple_task_card_pickup}
  \end{subfigure}
  \begin{subfigure}[t]{0.31\linewidth}\centering
    \includegraphics[trim={10 0 10 0}, clip, width=.99\linewidth]{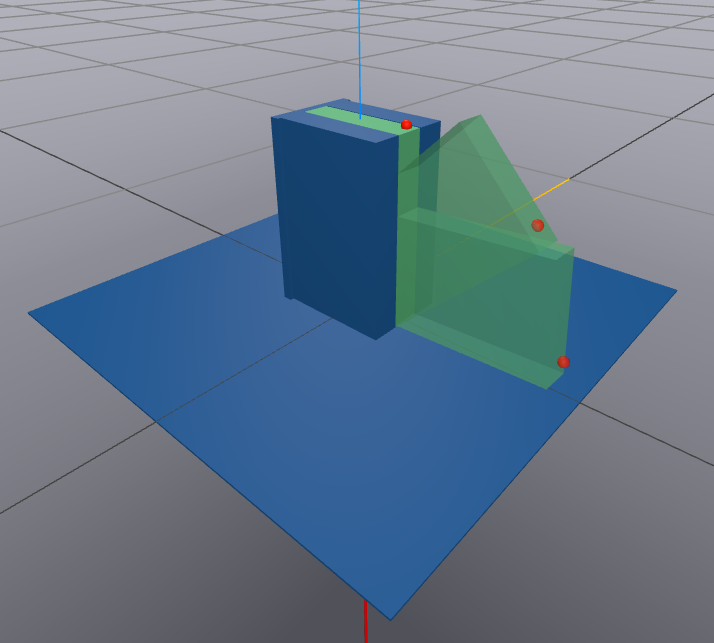}
    \caption{Book Removal}\label{fig:simple_task_bookshelf}
  \end{subfigure}
  \par\smallskip 
  \begin{subfigure}[t]{0.31\linewidth}\centering
    \includegraphics[trim={0 0 0 5}, clip, width=.99\linewidth]{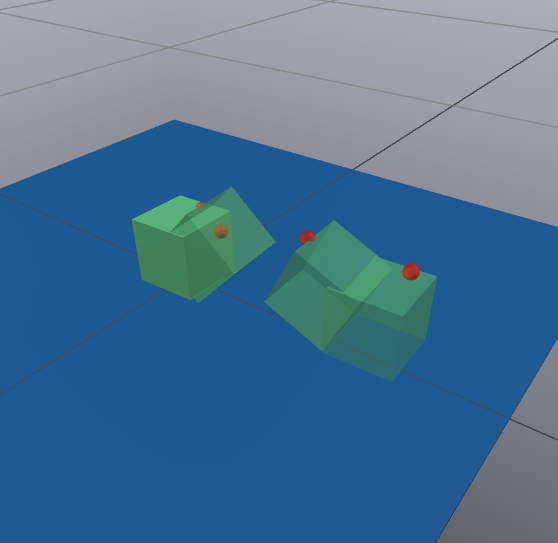}
    \caption{Box Flip}\label{fig:simple_task_box_flip}
  \end{subfigure}
  \begin{subfigure}[t]{0.31\linewidth}\centering
    \includegraphics[trim={2 0 2 0}, clip, width=.99\linewidth]{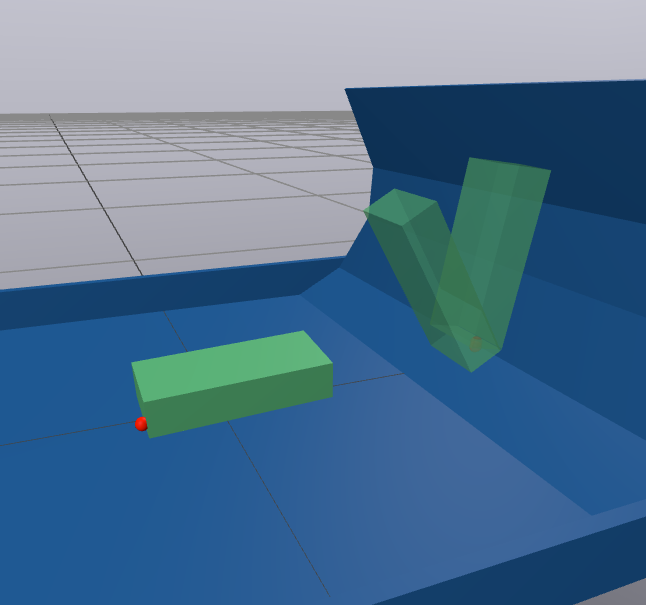}
    \caption{Half-pipe Flip}\label{fig:simple_task_ramp_flip}
  \end{subfigure}
  \begin{subfigure}[t]{0.31\linewidth}\centering
    \includegraphics[width=.99\linewidth]{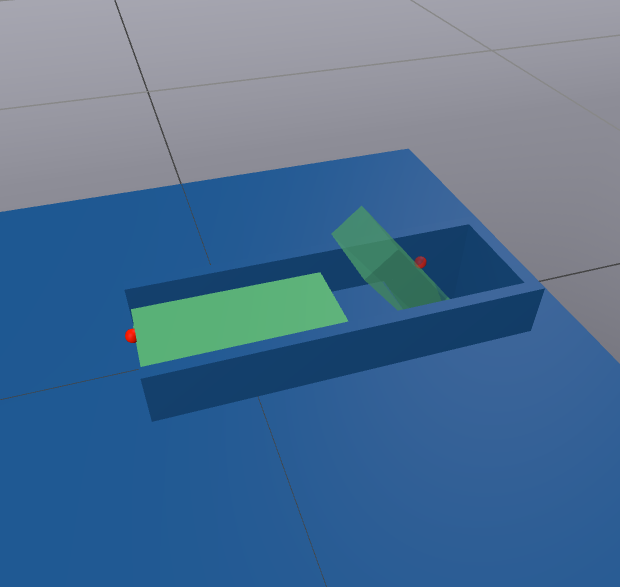}
    \caption{Prism Slot}
    \label{fig:simple_task_prism_slot}
  \end{subfigure}
\caption{Short-Horizon Extrinsic Dexterity Tasks}
\label{figs:simple_extrinsic_dexterity_tasks}
\end{figure}

We first consider single fingertip variants of the extrinsic dexterity tasks proposed in \cite{cheng2024mcts} (picking up a card, removing a book from a bookshelf, flipping a box) alongside some additional tasks (planar pushing, flipping via a half-pipe, sliding a prism into a slot), see Fig. \ref{figs:simple_extrinsic_dexterity_tasks}. To pick up the card with a single fingertip, we added a small lip in the environment for the finger to push the card against. In all cases, the finger leverages interaction with the environment to achieve its goal - for instance, in the half-pipe scenario, the finger must slide the prism up the half-pipe to let gravity tip it over backwards.

\setlength{\belowcaptionskip}{0pt}
\vspace{0.1cm}

\setlength{\tabcolsep}{5pt}
\begin{table}[h]
\centering
\begin{tabular}{llcccccc}
\toprule
\textbf{Method} & \textbf{Metric} & PP & CF & BR & BF & HP & PS \\
\midrule
\multirow{4}{*}{Ours}
 & Success       & 1.0  & 1.0  & 1.0  & 1.0  & 1.0  & 1.0  \\
 & Time [s]           & 6.04 & 2.06 & 0.75 & 3.02 & 3.28 & 2.56 \\
 & Modes      & 2.6  & 4.7  & 2.5  & 6.1  & 6.7  & 6.0  \\
 & Branches   & 73.5 & 60.1 & 11.7 & 82.4 & 90.5 & 54.7 \\
\midrule
\multirow{2}{*}{Bsl. 1~\cite{pang2023rrt}}
 & Success & 0.9 & 0.05       & 0.0      & 0.25  & $-$ & $-$ \\
 & Time [s]     & 3.1 & 17.9       & $>$30.3   & 10.3 & $-$ & $-$ \\
\midrule
\multirow{2}{*}{Bsl. 2~\cite{cheng2024mcts}}
 & Success & $-$ & 1.0 & 1.0 & 1.0 & $-$ & $-$ \\
 & Time [s]     & $-$ & 5.1 & 1.2 & 4.6 & $-$ & $-$ \\
\bottomrule
\end{tabular}
\caption{Performance on Short-Horizon Tasks}
\label{table:simple_performance_results}
\end{table}

We present the results of this experiment in Table~\ref{table:simple_performance_results}. 
The table columns correspond to each task: planar pushing (PP), card flipping (CF), book removal (BR), box flipping (BF), half-pipe (HP), prism slot (PS). 
The table rows correspond to each metric: Success is a binary value indicating that the goal configuration is reached, Time is the wall clock time until an exit condition is reached, Modes is the final plan's number of fingertip contact mode changes, and Branches is an indication of the search tree size. We collect statistics from running each experiment 20 times with different random seeds and report the average across trials.

We compare against two state of the art baselines for contact rich global planning from Pang et al. \cite{pang2023rrt} (Baseline 1) and Cheng et al. \cite{cheng2024mcts} (Baseline 2), using their open-source code. We re-implement the first four tasks for Baseline 1, and indicate the results reported in \cite{cheng2024mcts} for Baseline 2.

\begin{figure}
  \begin{subfigure}[t]{0.49\linewidth}\centering
    \includegraphics[width=.9\linewidth]{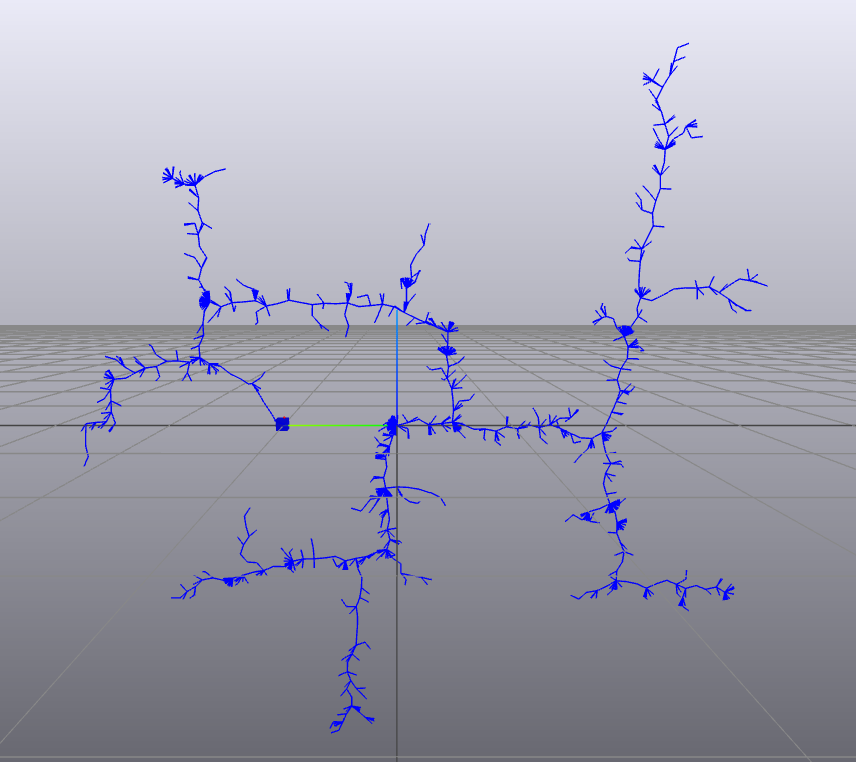}
    \caption{2D Planar Pusher}\label{figs:pang_baseline_2d}
  \end{subfigure}
  \begin{subfigure}[t]{0.49\linewidth}\centering
    \includegraphics[trim={0 0 0 0}, clip, width=.9\linewidth]{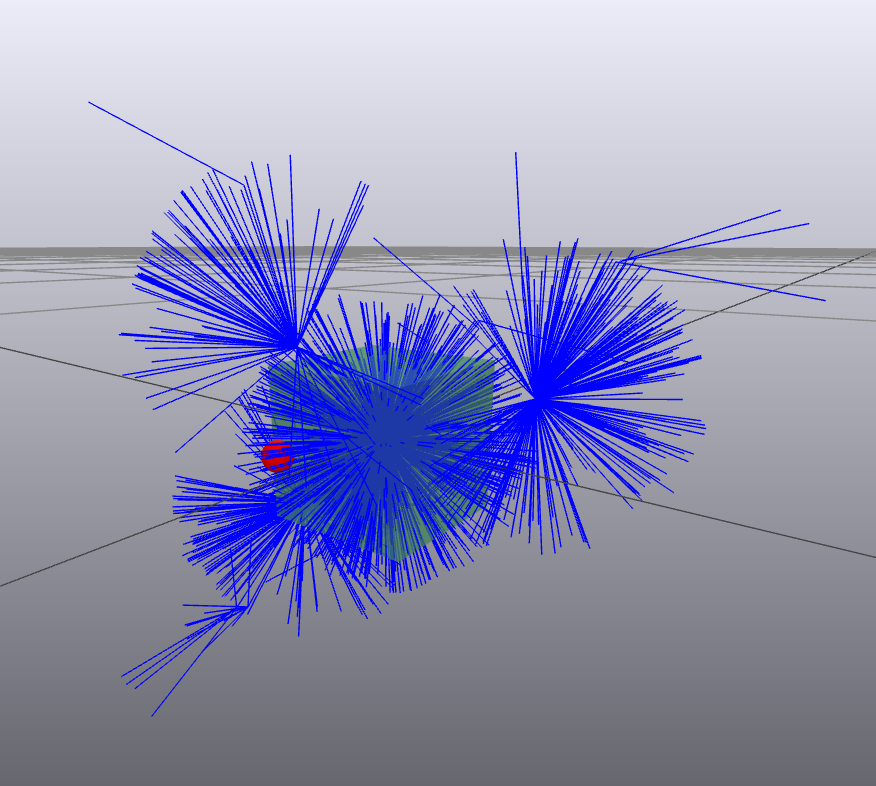}
    \caption{3D Rectilinear Pusher}
    \label{figs:pang_baseline_3d}
  \end{subfigure}
\caption{Plots of tree expansion using Baseline 1~\cite{pang2023rrt} for 2D and 3D pusher tasks. Tree branches shown in blue. 3D tasks collapse to breadth first search.
}
\label{figs:pang_baseline_expansion_comparison}
\end{figure}

Despite being successful on 2D Planar Pushing, the baseline from~\cite{pang2023rrt} struggles with short horizon extrinsic dexterity tasks for 3D objects because the tree search collapses to a breadth first search regime, see Fig.~\ref{figs:pang_baseline_expansion_comparison}. 
We suspect this is because their method relies on a distance metric that becomes poorly conditioned in higher dimensions: we sampled the distance metric over thousands of states and found that the eccentricity of their distance metric is above 0.9 (i.e. highly anisotropic) in 99\% of the samples  in 3D, while only 1\% are above 0.9 in 2D.
Our method avoids this challenge because our exploration is guided by the spectral decomposition. 

Our results indicate that we achieve comparable timing to the $\sim$5 seconds reported in the original experiments from~\cite{cheng2024mcts}. However, whereas their method separates the process of proposing object motions from the process of determining motion actuation via contact, our method jointly proposes distinct, actuatable motions without the need for nested tree search. This allows our method to scale to long-horizon tasks, which we present next. 

\setlength{\belowcaptionskip}{0pt}

\subsection{Long-horizon Extrinsic Dexterity Maze Task}

\begin{figure*}
\centering
\begin{subfigure}{.33\textwidth}
  \centering
  \includegraphics[width=.95\linewidth]{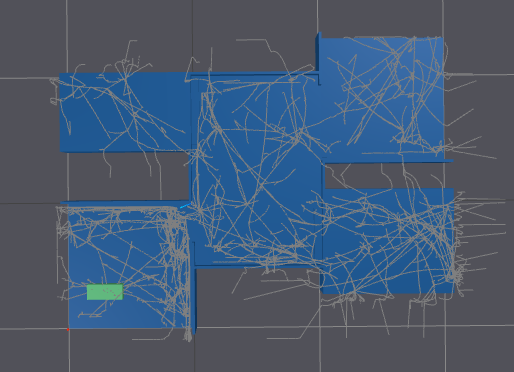}
  \caption{Our Method}
  \label{figs:ablation_full}
\end{subfigure}%
\begin{subfigure}{.33\textwidth}
  \centering
  \includegraphics[width=.95\linewidth]{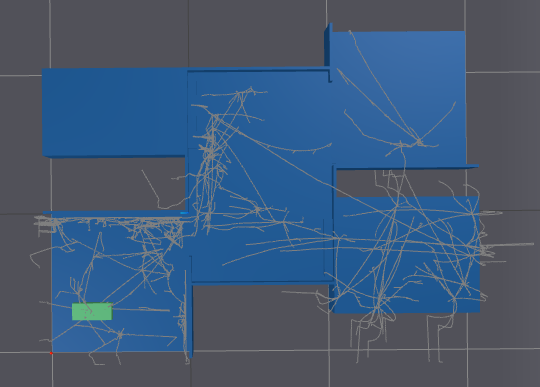}
  \caption{Ablation 1}
  \label{figs:ablation_1}
\end{subfigure}%
\begin{subfigure}{.33\textwidth}
  \centering
  \includegraphics[width=.95\linewidth]{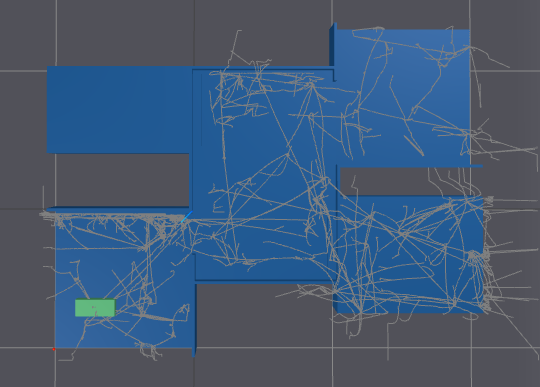}
  \caption{Ablation 2}
  \label{figs:ablation_2}
\end{subfigure}
\begin{subfigure}{.33\textwidth}
  \centering
  \includegraphics[width=.95\linewidth]{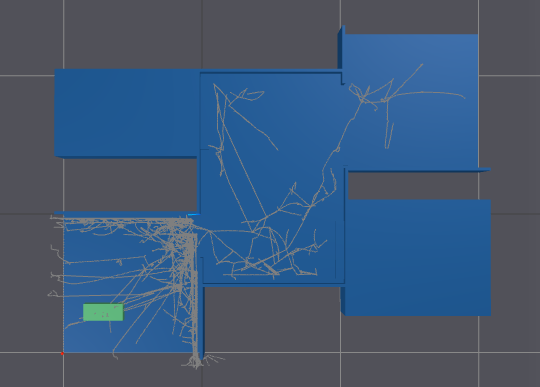}
  \caption{Ablation 3}
  \label{figs:ablation_3}
\end{subfigure}%
\begin{subfigure}{.33\textwidth}
  \centering
  \includegraphics[width=.95\linewidth]{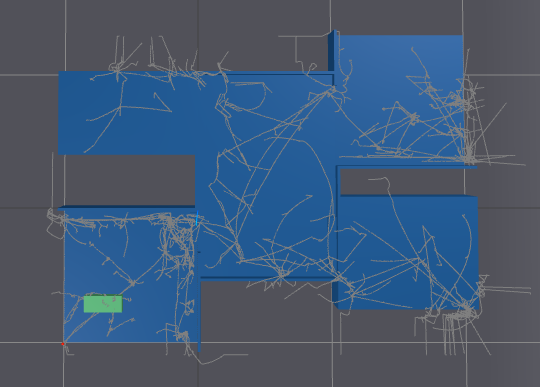}
  \caption{Ablation 4}
  \label{figs:ablation_4}
\end{subfigure}%
\caption{Plots of tree expansion after 60s of computation in the 3D maze environment}
\label{figs:ablations_tree_expansion}
\end{figure*}

To determine our methods's function and computational timing on a longer horizon problem, we consider a hand-crafted maze-environment formed by 5 shelves of varying heights (see Figs. \ref{figs:intropicture} and \ref{figs:ablations_tree_expansion}). This environment has rises, drops, protruding boxes, walls, and doorways for the object to move around and through. The 5 shelves are arranged in an X-shaped configuration, and we randomly sample each of the four non-central shelves as starting and ending locations for the planner. We run 50 random trials and report the average results in Table \ref{table:ablation_results}, where metrics other than Success Rate are only calculated for the trials that are successful. Trials are given 180 seconds of computer time to find a solution.
\setlength{\belowcaptionskip}{9pt}
\setlength{\belowcaptionskip}{-5pt}

Additionally, we ablate key elements of our algorithm to  determine their influence on performance. 
We also include Baseline 2 since it was successful on the short-horizon tasks. This baseline can fail in less than 180 seconds since there is a computational limit on the tree size, which we made as large as possible. We reported the average failure time as a lower bound on the solution time in Table \ref{table:ablation_results}.  

\begin{itemize}
\item \textit{Ablation 1}: During RRT Expansion, expand every trajectory from $\mathcal{A}(\mathbf{s}^*)$ instead of selecting a random one.
\item \textit{Ablation 2}: Remove the $\textsc{Kmeans}$ clustering step.
\item \textit{Ablation 3}: Remove the $\textsc{FilterProposals}$ step.
\item \textit{Ablation 4}: During construction of $\mathcal{A}(\mathbf{s}^*)$, replace the proposed $\mathbf{v}_i$ from Equation \ref{equations:eigen} with a random vector.
\item \textit{Baseline 2}: Solving the task using ~\cite{cheng2024mcts}.
\end{itemize}

A qualitative view of tree expansion under each ablation can be seen in Fig.  \ref{figs:ablations_tree_expansion}, which illustrates the discovered tree paths for the maze problem beginning on the bottom left shelf. This is a challenging initial condition since progress can only be made by tipping the object onto a box near the doorway to the central shelf.
Non-terminal trajectories were not available for plotting from Baseline 2.

\begingroup
\renewcommand{\arraystretch}{1.2}
\begin{table}[h]
\centering
\begin{tabular}{lccccc}
\toprule
 \makecell{\textbf{Method}} & \makecell{Success} & \makecell{Time [s]} &
   \makecell{Modes} & 
   \makecell{Branches} &
   \makecell{Length [m]} \\
\midrule
Ours   & 1.0  & 15.0  & 9.34 & 436 & 3.90 \\
Abl. 1 & 1.0  & 35.1  & 13.8 & 389 & 4.46 \\
Abl. 2 & 0.96 & 20.55 & 10.8 & 492 & 3.32 \\
Abl. 3 & 0.74 & 20.0  & 11.9 & 591 & 3.04 \\
Abl. 4 & 0.88 & 13.3  & 31.1 & 634 & 3.15 \\
Bsl. 2$^*$ & 0.0 & $>$32.5 & $\sim$ & $\sim$ & $\sim$ \\
\bottomrule
\end{tabular}
\caption{Ablation (Abl.) and Baseline (Bsl.) Performance. $^*$Ablations 1–4 have 50 trials and Baseline 2 has 10 trials.}
\label{table:ablation_results}
\end{table}
\endgroup

As shown in Table~\ref{table:ablation_results}, the algorithm's perfect success rate is indicative of its ability to plan through multiple contact modes across long task horizons. 
This success happens in spite  of the fact that many motions plans must pass through very specific sequences of object motions in order to be successful, or through small gaps, such as when going from the bottom-left and upper-right platforms to the central platform. The tree efficiently considers a wide variety of diverse possible motions in order to quickly cover the entire area, as evidenced by the coverage of the maze in Fig. \ref{figs:ablation_full}. 

The results in Table~\ref{table:ablation_results} also demonstrate that the wall clock planning time is less than the generated plan's execution time, assuming that the plans are followed with an object speed of 0.05-0.2 m/s (depending upon the local complexity along the planned trajectory). These long-horizon motions take upwards of 30-45 seconds to execute. 
This means that future plans can begin to be generated during the original object motion, allowing real-time re-planning.

The relative performance of our algorithm and Baseline 2 clearly demonstrates improvements over existing work~\cite{cheng2024mcts}. 
We first associate this performance gap with issues in dynamic feasibility of the exploration: whereas~\cite{cheng2024mcts} proposes object motions before determining whether they can be achieved via contact, our proposed motion uses the range space of $\mathbf{A}(q)$ to inform whether the fingertip is capable of actuating those motions. 
Secondly, their nested search structure coupled with the long-horizon duration limits the number of available rollouts to build the search tree. 

The remaining ablations knock out key subroutines in our algorithm and reveal our method's most critical features. 
Ablation 1 favors a breadth first exploration by expanding all the children of a node before continuing the exploration process. 
Although this ablation leads to a smaller average number of branches to reach a solution, it requires more total computation time. 
The timing gap with Ablation 2 demonstrates that planning duration and tree expansion efficiency arises from clustering the proposed actions. This result is expected given that clustering reduces the number of exploratory choices to a few effective ones. 
The performance gap with Ablation 3 demonstrates the importance of proposal filtering, which prevents non-physical behavior like pulling on objects or pushing into the solid surroundings. 
Ablation 4 follows random vectors rather than informed principle directions, resulting in lower success rate, higher number of contacts, and higher number of explored branches. 

Finally, we present a visual analysis of the methods and ablations in Fig.~\ref{figs:ablations_tree_expansion}. 
We notice that our full method is able to cover the entire space within the allotted time frame, while most of the ablations fail to find at least one of the platforms. 
Ablation 4's relatively high success rate demonstrates that exploratory rollouts in \textit{any} direction in a high fidelity simulator form a key part of reaching the goal. However, the shorter choppier motion segments and increased contact changes that come from ignoring the inverse dynamics and dynamically actuatable motions cover the space less thoroughly and are likely far less robust in deployment.

\section{Conclusion and Future Work}

This work proposed a model-based planning algorithm that is specialized for manipulation tasks involving long-horizons and multiple contact modes requiring physical interaction with the environment. 
The core of our proposed method is a new hierarchical decomposition: tree search over trajectories generated from the spectrum of the inverse dynamics. 
This combination inherits global exploration from search and efficient exploration from local information in the inverse dynamics equation. 
The method's effectiveness was demonstrated in high-fidelity simulations of long-horizon tasks, where our method generated complete solutions faster than state-of-the-art baselines and achieved a perfect success rate. 
Our simulations also demonstrated planning times that are less than plan execution time, demonstrating a capability for real-time deployment and online re-planning.
In future work, we will continue to develop this decomposition with experimental demonstrations and algorithmic improvements focusing on transient dynamic contact motions and robustness to model uncertainty. 

\printbibliography

@misc{howell2022predictivesamplingrealtimebehaviour,
      title={Predictive Sampling: Real-time Behaviour Synthesis with MuJoCo}, 
      author={Taylor Howell and Nimrod Gileadi and Saran Tunyasuvunakool and Kevin Zakka and Tom Erez and Yuval Tassa},
      year={2022},
      eprint={2212.00541},
      archivePrefix={arXiv},
      primaryClass={cs.RO},
      url={https://arxiv.org/abs/2212.00541}, 
}

@inproceedings{toussaint2018tamp,
  title     = {Differentiable Physics and Stable Modes for Tool-Use and Manipulation Planning - Extended Abtract},
  author    = {Toussaint, Marc and Allen, Kelsey R. and Smith, Kevin A. and Tenenbaum, Joshua B.},
  booktitle = {Proceedings of the Twenty-Eighth International Joint Conference on
               Artificial Intelligence, {IJCAI-19}},
  publisher = {International Joint Conferences on Artificial Intelligence Organization},
  pages     = {6231--6235},
  year      = {2019},
  month     = {7},
  doi       = {10.24963/ijcai.2019/869},
  url       = {https://doi.org/10.24963/ijcai.2019/869},
}

@ARTICLE{kingston2023scalingmultimodalplanning,
  author={Kingston, Zachary and Kavraki, Lydia E.},
  journal={IEEE Transactions on Robotics}, 
  title={Scaling Multimodal Planning: Using Experience and Informing Discrete Search}, 
  year={2023},
  volume={39},
  number={1},
  pages={128-146},
  doi={10.1109/TRO.2022.3197080}}

@book{feedbackcontrolhybridsystems,
author={Eric R. Westervelt
and Jessy W. Grizzle
and Christine Chevallereau
and Jun-Ho Choi
and Benjamin Morris},
title={Feedback Control of Dynamic Bipedal Robot Locomotion},
year={2007},
}

@article{riviere2021neural,
  title={Neural tree expansion for multi-robot planning in non-cooperative environments},
  author={Riviere, Benjamin and H{\"o}nig, Wolfgang and Anderson, Matthew and Chung, Soon-Jo},
  journal={IEEE Robotics and Automation Letters},
  volume={6},
  number={4},
  pages={6868--6875},
  year={2021},
  publisher={IEEE}
}

@inproceedings{zhu2023efficient,
  title={Efficient object manipulation planning with monte carlo tree search},
  author={Zhu, Huaijiang and Meduri, Avadesh and Righetti, Ludovic},
  booktitle={IEEE/RSJ Int. Conf.  intelligent robots and systems },
  pages={10628--10635},
  year={2023}
}

@article{morgan2014model,
  title={Model predictive control of swarms of spacecraft using sequential convex programming},
  author={Morgan, Daniel and Chung, Soon-Jo and Hadaegh, Fred Y},
  journal={J. Guidance, Control, and Dynamics},
  volume={37},
  number={6},
  pages={1725--1740},
  year={2014}
}

@inproceedings{marcucci2019mipaffinesystems,
author = {Marcucci, Tobia and Tedrake, Russ},
year = {2019},
month = {04},
pages = {},
title = {Mixed-Integer Formulations for Optimal Control of Piecewise-Affine Systems},
doi = {10.1145/3302504.3311801}
}

@article{marcucci2020warmstartmip,
author = {Marcucci, Tobia and Tedrake, Russ},
year = {2020},
month = {07},
pages = {1-1},
title = {Warm Start of Mixed-Integer Programs for Model Predictive Control of Hybrid Systems},
volume = {PP},
journal = {IEEE Trans. on Automatic Control},
doi = {10.1109/TAC.2020.3007688}
}

@misc{drake,
 author = "Russ Tedrake and the Drake Development Team",
 title = "Drake: Model-based design and verification for robotics",
 year = 2019,
 url = "https://drake.mit.edu"
}

@article{huntcrossley,
  author={K. Hunt and F. Crossley},
  title={Coefficient of restitution interpreted as damping in vibroimpact},
  journal={J. of Applied Mechanics},
  year={1975},
  volume={42},
  pages={440-445},
  number={2} 
}

@article{
riviere2024sets,
author = {Benjamin Rivière  and John Lathrop  and Soon-Jo Chung },
title = {Monte Carlo tree search with spectral expansion for planning with dynamical systems},
journal = {Science Robotics},
volume = {9},
number = {97},
pages = {eado1010},
year = {2024},
doi = {10.1126/scirobotics.ado1010},
URL = {https://www.science.org/doi/abs/10.1126/scirobotics.ado1010}
}

@ARTICLE{lecleach2024cimpc,
  author={Le Cleac'h, Simon and Howell, Taylor A. and Yang, Shuo and Lee, Chi-Yen and Zhang, John and Bishop, Arun and Schwager, Mac and Manchester, Zachary},
  journal={IEEE Trans. on Robotics}, 
  title={Fast Contact-Implicit Model Predictive Control}, 
  year={2024},
  volume={40},
  number={},
  pages={1617-1629},
  doi={10.1109/TRO.2024.3351554}}

@INPROCEEDINGS{ames2016onlinegaits,
  author={Hereid, Ayonga and Kolathaya, Shishir and Ames, Aaron D.},
  booktitle={IEEE Conf. Decision and Control}, 
  title={Online optimal gait generation for bipedal walking robots using legendre pseudospectral optimization}, 
  year={2016},
  volume={},
  number={},
  pages={6173-6179},
  doi={10.1109/CDC.2016.7799218}}

@ARTICLE{scaramuzza2024drones,
  author={Hanover, Drew and Loquercio, Antonio and Bauersfeld, Leonard and Romero, Angel and Penicka, Robert and Song, Yunlong and Cioffi, Giovanni and Kaufmann, Elia and Scaramuzza, Davide},
  journal={IEEE Trans. on Robotics}, 
  title={Autonomous Drone Racing: A Survey}, 
  year={2024},
  volume={40},
  number={},
  pages={3044-3067},
  doi={10.1109/TRO.2024.3400838}}

@misc{kurtz2025generativepredictivecontrolflow,
      title={Generative Predictive Control: Flow Matching Policies for Dynamic and Difficult-to-Demonstrate Tasks}, 
      author={Vince Kurtz and Joel W. Burdick},
      year={2025},
      eprint={2502.13406},
      archivePrefix={arXiv},
      primaryClass={cs.RO},
      url={https://arxiv.org/abs/2502.13406}, 
}

@ARTICLE{zhu2025learningfrommodels,
  author={Zhu, Huaijiang and Zhao, Tong and Ni, Xinpei and Wang, Jiuguang and Fang, Kuan and Righetti, Ludovic and Pang, Tao},
  journal={IEEE Robotics and Automation Letters}, 
  title={Should We Learn Contact-Rich Manipulation Policies From Sampling-Based Planners?}, 
  year={2025},
  volume={10},
  number={6},
  pages={6248-6255},
  doi={10.1109/LRA.2025.3564701}}

@article{pang2023rrt,
  doi = {10.48550/ARXIV.2206.10787},
  url = {https://arxiv.org/abs/2206.10787},
  author = {Pang, Tao and Suh, H. J. Terry and Yang, Lujie and Tedrake, Russ},
  title = {Global Planning for Contact-Rich Manipulation via Local Smoothing of Quasi-dynamic Contact Models},
  publisher = {arXiv},
  year = {2022},
  copyright = {Creative Commons Attribution 4.0 International}
}

@article{cheng2024mcts,
  author={Cheng, Xianyi and Patil, Sarvesh and Temel, Zeynep and Kroemer, Oliver and Mason, Matthew T.},
  journal={IEEE Robotics and Automation Letters}, 
  title={Enhancing Dexterity in Robotic Manipulation via Hierarchical Contact Exploration}, 
  year={2024},
  volume={9},
  number={1},
  pages={390-397},
  doi={10.1109/LRA.2023.3333699}}

@article{posa2014cito,
author = {Michael Posa and Cecilia Cantu and Russ Tedrake},
title ={A direct method for trajectory optimization of rigid bodies through contact},
journal = {Int. J. Robotics Research},
volume = {33},
number = {1},
pages = {69-81},
year = {2014},
doi = {10.1177/0278364913506757}
}

@misc{suh2025dexterouscontactrichmanipulationcontact,
      title={Dexterous Contact-Rich Manipulation via the Contact Trust Region}, 
      author={H. J. Terry Suh and Tao Pang and Tong Zhao and Russ Tedrake},
      year={2025},
      eprint={2505.02291},
      archivePrefix={arXiv},
      primaryClass={cs.RO},
      url={https://arxiv.org/abs/2505.02291}, 
}

@misc{pizero2024,
  title         = {$\pi^0$: A Vision--Language--Action Flow Model for General Robot Control},
  author        = {Black, Kevin and Brown, Noah and Driess, Danny and Esmail, Adnan and Equi, Michael and Finn, Chelsea and Fusai, Niccolo and Groom, Lachy and Hausman, Karol and Ichter, Brian and Jakubczak, Szymon and Jones, Tim and Ke, Liyiming and Levine, Sergey and Li-Bell, Adrian and Mothukuri, Mohith and Nair, Suraj and Pertsch, Karl and Shi, Lucy Xiaoyang and Tanner, James and Vuong, Quan and Walling, Anna and Wang, Haohuan and Zhilinsky, Ury},
  year          = {2024},
  eprint        = {2410.24164},
  archivePrefix = {arXiv},
  primaryClass  = {cs.RO},
  url           = {https://arxiv.org/abs/2410.24164}
}

@article{lavalle1998rapidly,
  title={Rapidly-exploring random trees: A new tool for path planning},
  author={LaValle, Steven},
  journal={Research Report 9811},
  year={1998},
  publisher={Department of Computer Science, Iowa State University}
}

@article{karaman2011sampling,
  title={Sampling-based algorithms for optimal motion planning},
  author={Karaman, Sertac and Frazzoli, Emilio},
  journal={Int. J.  Robotics Research},
  volume={30},
  number={7},
  pages={846--894},
  year={2011},
  publisher={Sage Publications Sage UK: London, England}
}

@misc{levine2016endtoendtrainingdeepvisuomotor,
      title={End-to-End Training of Deep Visuomotor Policies}, 
      author={Sergey Levine and Chelsea Finn and Trevor Darrell and Pieter Abbeel},
      year={2016},
      journal={Journal of Machine Learning Research},
      volume={17}
}

@article{kurtz2023inverse,
  author = {Vince Kurtz and Alejandro Castro and Aykut Özgün Önol and Hai Lin},
title ={Inverse dynamics trajectory optimization for contact-implicit model predictive control},

journal = {Int. J. Robotics Research},
year = {2025},
doi = {10.1177/02783649251344635},
}

\end{document}